\documentclass{article}
\usepackage{ijcai18}

\usepackage{times}
\usepackage{xcolor}
\usepackage{soul}
\usepackage[utf8]{inputenc}
\usepackage[small]{caption}
\usepackage{amsmath,amssymb}

\usepackage{diagbox}
\usepackage{graphicx}
\usepackage{subcaption}
\graphicspath{ {images/} }

\usepackage{array,graphicx}
\usepackage{booktabs}
\usepackage{pifont}
\usepackage{siunitx}
\usepackage{makecell}
\usepackage{comment}
\usepackage{hyperref}

\title{Formula RL: Deep Reinforcement Learning for Autonomous Racing using Telemetry Data}
\author{
    Adrian Remonda$^1$, 
    Sarah Krebs$^1$, 
    Eduardo Veas$^1$, 
    Granit Luzhnica$^1$, 
    Roman Kern$^1$
    \\ 
    $^1$ Know-Center - Graz, Austria \\
    \{aremonda, skrebs, eveas, gluzhnica, rkern\} @ know-center.at
}
\usepackage{amsmath}
\usepackage{amsfonts}
\usepackage{subcaption}
\usepackage{graphicx}
\usepackage{varwidth}
\usepackage{booktabs}
\usepackage{multirow}
\usepackage{algpseudocode}% http://ctan.org/pkg/algorithmicx
\usepackage{amsmath} % for 'cases' env.

\newif\ifcheat
\cheattrue
\ifcheat
	\def\baselinestretch{0.97}
\fi

\graphicspath{ {images/} }

\begin{document}

\maketitle

\def\toprule{\hline\hline}%
\def\bottomrule{\hline\hline}%
\def\midrule{\hline}%

\begin{abstract}
This paper explores the use of reinforcement learning (RL) models for autonomous racing.
%This paper puts reinforcement learning (RL) models to race literally.
%We focus on a specific part of the autonomous driving problem: the racing car. 
In contrast to passenger cars, where safety is the top priority, a racing car aims to minimize the lap-time. We frame the problem as a reinforcement learning task with a multidimensional input consisting of the vehicle telemetry, and a continuous action space.
To find out which RL methods better solve the problem and whether the obtained models generalize to driving on unknown tracks, we put 10 variants of deep deterministic policy gradient (DDPG) to race in two experiments: i)~studying how RL methods learn to drive a racing car and ii)~studying how the learning scenario influences the capability of the models to generalize. 
Our studies show that models trained with RL are not only able to drive faster than the baseline open source handcrafted bots but also generalize to unknown tracks.
\end{abstract}

\section{Introduction}
%Autonomous driving has received a lot of interest from the media and research alike, due to its potential to change how mobility and transport may look like in the future. Our research is concerned with driving an autonomous car at its physical limits. This may apply to situations like avoiding obstacles that suddenly appear, or, like in our case, driving a racing car. Here, the objective is to drive the car around a racing track so as to achieve the lowest possible lap-time and to obtain an algorithm that is as generalizable as possible to assist professional drivers to improve their racing line. Hence, the car has to move at its physical limits. In common practice, this problem is addressed by applying methods from the field of control theory. These methods utilize heuristics and demand domain knowledge to manually tune the model's parameters ~\cite{Segers2014}. But each racing track has its own peculiar challenges and effectiveness on one track does not necessarily translate to other tracks: these methods often require a different set of heuristics and parameters for each new situation.

Autonomous driving has received a lot of interest from the media and research alike, due to its potential to change how mobility and transport may look like in the future. However, within the domain of autonomous driving, there are different scenarios which dictate the objective for which to optimize. The focus of our research is to provide autonomous driving for racing cars in order to assist professional drivers to improve their racing line.
In this case, the ultimate goal of our driver model is to drive the car around a racing track so as to achieve the lowest possible lap-time, preferably by reaching the physical limits of the car. In common practice, this problem is addressed by applying methods from the field of control theory. These methods utilize heuristics and demand domain knowledge to tune the model's parameters~\cite{Segers2014} manually. As each racing track has its peculiar challenges, these methods often require a different set of heuristics and parameters for each new situation.

Reinforcement learning aims at training an agent to learn to interact with an environment such as to maximize some notion of long-term reward. 
Combining RL with deep learning, problems with high-dimensional state spaces can be solved~\cite{DBLP:journals/corr/MnihKSGAWR13,mnih2015humanlevel}. With algorithms like deep deterministic policy gradient (DDPG), deep RL can be extended to allow for solving continuous action space optimization problems \cite{DBLP:journals/corr/LillicrapHPHETS15}. This is an essential prerequisite for our use case as the racing car expects inherently continuous control (i.e. steering, brake, and throttle).

As regards autonomous driving, a lot of attention has been devoted to image processing, to analyze information based on images from cameras~\cite{DeepDriving7410669,DBLP:articles/kt2013,DBLP:articles/Drews2017,DBLP:articles/Jaritz2018}. 
%Our ultimate goal is to train in simulation and then transfer to real-world scenario
In general, processing video input takes significant time and resources in training phases: from rendering to learning from them.
%However, racing car setups do not have cameras in place and extending such setups would require infrastructure changes which might not even be feasible. 
Instead, a physics engine for racing simulation often runs headless to produce telemetry data at high rates (1000Hz), and racing cars already have the infrastructure to capture and make use of telemetry data which could be utilised for our purpose. Thus, our goal in this paper is to investigate how well different RL models can perform by using solely telemetry data streams. 
Besides, images contain a wealth of information which makes it difficult to extract the essentials needed to drive a racing car to its physical limits. For instance, there are many corners were the visibility is limited by obstacles. That is why, human pilots need to learn the track by heart before the race which helps them to decide how to manage (enter or leave) corners. Similarly, our algorithms use a (partially observable) racing line as reference for decision. 
It is common knowledge that besides vision and knowledge of the track, drivers heavily rely on the lateral and longitudinal acceleration that they feel through the vestibular system as well as the saturation of the tires by sensing the torque present in the steering wheel~\cite{DBLP:articles/BentleyR98}. Such information is well represented by the telemetry data of racing cars.

%Our three research questions are: RQ1 -- \rqa? RQ2 -- \rqb? RQ3 -- \rqc? In the spirit of the competition, our experimental design is a Formula RL tournament, where 10 variants of DDPG contend in two studies. In the first study, 10 algorithms are trained to drive on a simple racing track. Algorithms producing the fastest models in terms of lap-time are promoted to a second part, which compares learning (re-trains algorithms) to drive on difficult tracks. In the second study the best models are tested on unknown tracks to assess how the current state-of-the-art in deep RL generalizes to unseen situations. 

Given our goal of autonomous racing, this work is driven by two research questions, with the first being (\textbf{RQ1}): \textit{is it feasible to learn a driver model that effectively drives a racing car by relying only on telemetry data as input? More concretely, which combination of algorithm/architecture is best suited to solve this task?}. Additionally, inspired by the common practice of human pilots, that train for the specific racing track (prior to the race), even though they already are proficient at driving, we explore the second research question (\textbf{RQ2}): \textit{how well such algorithms trained on one track are able to generalize to other tracks?}

In the spirit of the competition, our experimental design is a tournament, where 10 variants of DDPG compete with each other in two studies. In the first study, 10 algorithms are trained to drive on a simple racing track. Algorithms producing the fastest models in terms of lap-time are promoted to a second part, to compare learning to drive on difficult tracks. In the second study the best models are tested on unknown tracks to assess how the current state-of-the-art in deep RL generalizes to unseen situations. 

%To the best of our knowledge, we are the first to explore racing line optimization problem using RL and to beat any open source bot with an algorithm self-thought completely from scratch. Our contributions to RL algorithms are: i) a simple scheme that reduces the exploration space when two outputs should be mutually exclusive in continuous action spaces, ii) a long short term memory (LSTM) layer in the DDPG's critic network to allow for taking advantage of past experiences, iii) a modification of the target to solve an issue that arises when sampling end-of-episode transitions. In addition, we report iv) a contribution in experimental methodology. Research on RL usually reports learning curves (how the algorithm learns). We also benchmark the models resulting from different algorithms on the real task and in unseen environments (how the model performs). Our research clearly shows that these tests need to be considered in the racing domain. %Our work overlaps two areas of research: autonomous racing and reinforcement learning.

In our studies, RL models trained (from scratch) outperform the best performing open source bots available for our simulation environment. As a result, the faster models deliver a new racing line that leads to better performance. 
To the best of our knowledge, we are the first to explore racing line optimization problem using RL, which the main contribution of this work.
Our further contributions to the modeling the autonomous racing problem using RL methods include: a simple scheme that reduces the exploration space when two outputs should be mutually exclusive in continuous action spaces and a modification of the target to solve an issue that arises when sampling end-of-episode transitions. Additionally, we propose using the look ahead curvature (LAC) which provides information regarding the upcoming shape of the track recorded in a previous lap.
We also benchmark the models resulting from different algorithms on the trained and unseen environments (racing tracks). Our research clearly shows that these tests need to be considered in the racing domain. %Our work overlaps two areas of research: autonomous racing and reinforcement learning.
(See videos with results \url{https://www.youtube.com/watch?v=GRYqOgb6DJQ})

\section{Background}

\subsection{Autonomous Racing}
\label{sec:motorsports}

Autonomous driving has matured into a field where reliable models are needed for obstacle detection and avoidance, maneuver initiation and recovery, while driving is reduced to path planning and path following. These models require a sophisticated perception of the environment, which is why much autonomous driving research focuses on image processing. The context of motorsports has its own requirements, that significantly influence the definition of the problem space. In contrast to passenger vehicles, where safety is the top priority, the main objective of a racing vehicle is to minimize the lap-time. 
One way to do this is to find, within the boundaries of the track, the trajectory where the car can move with the best lap-time--\emph{the optimal racing line}. The optimal racing line is the best compromise between the shortest path and the trajectory that allows to achieve the highest speeds \cite{BRAGHIN20081503}. It depends on several factors including the track shape, the car aerodynamics, grip, etc. \cite{Cardamone:2010:SPRLA}. Hence, the problem of trajectory planning is a bounded optimization problem that requires to take into account not only the geometry of the track but also vehicle dynamics. 
Autonomous racing thus requires a perception of the vehicle dynamics in relation with the environment. Typically, the optimal racing line is calculated offline or is estimated by the reference of an expert human driver. 

Attempts to achieve the lowest possible lap-times with autonomous racing cars typically combine control theory, determining and utilizing the optimal racing line, and/or optimizing the driver model directly. \cite{Cardamone:2010:SPRLA} calculated the ideal racing line using a genetic algorithm and then measured the lap-time of a line-follower bot. Their method outperformed the previous state-of-the-art models by a small margin while ours outperformed the state-of-the-art by a large margin and without having the necessity of having a line follower bot and just a reference line (this is typically done in professional racing scenarios, where a loose racing line exists that racers have as a reference). Another disadvantage of their method, is that it is unable to generalize to different tracks and limited to the performance of the bot. The racing line needs to be re-calculated for every new track or new car. 

% Related work using images
The aim of~\cite{DBLP:articles/kt2013} was to demonstrate that a car can be autonomously driven by using images. Their approach performed a prepossessing step in order to reduce the state space by representing it in the frequency domain, converting the images in a set of coefficients that are then transformed into weight matrices via an inverse Fourier-type transform. However, their approach targets driving in general and not racing which would require to minimise the lap time. In contrary, we aim to reduce the lap time which is the most important factor in racing.
\cite{DBLP:articles/Sallab2017} tackle the generalization issues in traditional imitation learning this means that they need to have demonstrations in all tracks. This demonstrations came from a traditional proportional-integral-derivative controller (PID controller) with access to the position of the ego car with respect to the left and right lanes. All the previously mentioned papers try to follow the center of the track, which makes it impossible to achieve the optimal racing line. We are using as a loose hint a racing line that we then improve by a large margin. ~\cite{DBLP:articles/kt2013}, ~\cite{DBLP:articles/Jaritz2018} used discrete action space while we are using continuous actions space.  Our preliminary studies (not included in this paper) showed that for a one-dimensional action space (steering wheel), a discrete actions space algorithm such as DQN might learn a policy that is able to drive on the track, but is far from reaching the lap-times achieved by the continuous actions space algorithm DDPG (limited to steering wheel control as well). In our opinion,it is not possible to drive to the limits by discretizing the throttle and steer. In comparison, human gamers of racing games/sims consider a dedicated steering wheel a worthwhile investment. In professional simulators, as the ones used to train F1 pilots, it is a must. 
\cite{DBLP:articles/Drews2017} present a deep learning method to generate cost maps learned from human demonstrations. The cost map is then feed to a model predictive control algorithm (MPC) that runs in real time on a real 1:5 scale autonomous vehicle by sampling trajectories using a model of the dynamics of the car. The dynamics model was learned directly from the data. Although, this work represents the state-of-the-art in the field of control theory and it has several advantages it still needs labeled data that would have to be recorded for each different track. It also has the drawback that the driving performance is limited by the quality of the human demonstrations.

\subsection{Motorsports}
\label{sec:motorsports}
As above introduced, the optimal racing line depends on factors including the track shape, the car aerodynamics, the grip, etc. \cite{Cardamone:2010:SPRLA}. It is often calculated offline or estimated by the reference of an expert human driver.

\paragraph{Racing line.}

\begin{figure}
\centering
  \includegraphics[width=0.30\columnwidth, height=3cm]{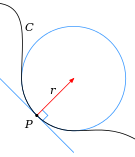}
  \includegraphics[width=0.40\columnwidth,height=3cm]{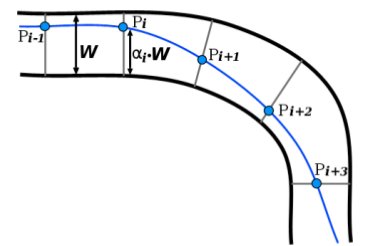}
  \caption[]{\emph{Left:} Geometrical representation of the radious of curvature \emph{Right:} Racing line representation \cite{Cardamone:2010:SPRLA} }
  \label{fig:racingline}
\end{figure}

The racing line is defined as a sequence of points on the track. Each point $P_i$ of the racing line can be represented by the pair $\langle \delta_i, \alpha_i \rangle$. As depicted in Figure~\ref{fig:racingline}, $\alpha_i$ is the lateral distance of the point from one of the track borders (e.g., from the right border), normalized by the track width $W$, and $\delta_i$ the distance of the point from the track starting line, computed along the track axis.

%\paragraph{Curvature.}
Figure~\ref{fig:racingline} shows the \textbf{curvature} $\kappa$ of the curve $C$ at a given point P in track is defined by the inverse of the curvature radius ($r$) at that point, and it's given by $\kappa = \frac{\partial \theta}{\partial s}$ where $\kappa$ is the curvature at a segment $\partial s$ with a change of angle  $\partial \theta$\footnote{https://en.wikipedia.org/wiki/Curvature}. The smaller the curvature, the higher the speed that a car can maintain along the racing line. The maximum possible speed without loosing grip is given by $v_{max} = \sqrt{\mu\rho (g + F_a/m)}$
%\begin{equation}\label{eq:curvature}
%v_{max} = \sqrt{\mu\rho \Bigg(g + \dfrac{F_a}{m}\Bigg)},
%v_{max} = \sqrt{\mu\rho (g + F_a/m)},
%\end{equation}
where $m$ is the car's mass, $\mu$ is the tire-road friction coefficient, $\rho$ is the curvature radius and $F_a$ is the downforce.
The shortest path is the sequence of points in one lap that results on the shortest distance, while the minimum curvature path is the sequence of points that allows to complete one lap with the minimum possible curvature.  \cite{Cardamone:2010:SPRLA}. 
%The curvature plays a major role in calculating the optimal racing line
%The systematic evaluation and comparison of algorithms is important to provide baselines for other researchers, but also to gain our understanding of the strengths and limitations of these algorithms.

\subsection{Reinforcement Learning}

Reinforcement learning deals with the problem of learning optimal behaviours for the interaction of an agent with an environment by trial and error, such as to maximize the accumulated reward obtained from the environment. It is assumed that the interaction of the agent with the environment takes place in discrete time steps \(t\). At each step, starting from a state \(s_t\), the agent executes an action \(a_t\) and receives a reward \(r_t\) and a new state \(s_{t+1}\) from the environment. 
The return from a state is defined as the sum of discounted future reward, \(R_t = \sum_{i=t}^T \gamma^{(i-t)} r_i\), where \(\gamma\in(0,1]\) is a discount factor and \(T\) is a terminal time step after which the process restarts. The objective of RL is to learn a policy \(\pi\), mapping states to actions, that maximizes the return from the start distribution. 
There are two main approaches for solving RL problems: methods based on value functions and policy search. So called actor-critic approaches employ both value functions and policy search.
A policy \(\pi\) defines the agent’s behavior by mapping states to actions. 
A value function provides an estimation of the future return and thus can be used to evaluate how good an action or state is. 
An action value function \(Q^\pi(a_t,s_t) = \mathbb{E}_\pi[R_t|s_t, a_t]\) estimates the return starting from state \(s_t\), taking action \(a_t\), and then following policy \(\pi\).

In deep RL the algorithm components are implemented as deep neural networks. 
%which allows for scaling to previously intractable, high-dimensional problems. 
The first successful deep RL algorithm was deep Q-network (DQN)\cite{DBLP:journals/corr/MnihKSGAWR13}, which 
%achieved human level performance on many Atari video games by learning directly from pixels. DQN and its extensions 
succeeded at solving problems with high-dimensional state spaces (e.g. pixels), but can only handle discrete, low-dimensional action spaces (e.g. left, right). Driving a car requires continuous actions (steering, throttle). The algorithms thereto are topic of the next section.

\section{Algorithms}
\label{sec:algos}

\subsection{Deep Deterministic Policy Gradient (DDPG)}

By combining the insights of DQN with the actor-critic deterministic policy gradient algorithm, DDPG \cite{DBLP:journals/corr/LillicrapHPHETS15} allows for solving a wide variety of continuous control tasks. DDPG utilizes an actor function \(\mu(s|\theta^\mu)\), specifying the current policy, and a critic function \(Q(s,a|\theta^Q)\), both approximated by neural networks. At each step, based on the current state \(s_t\), the agent chooses an action according to \(a_t = \mu(s_t|\theta^\mu) + \mathcal{N}\), with a noise process \(\mathcal{N}\) to allow for exploration, and obtains a reward \(r_t\) and a new state \(s_{t+1}\) from the environment. The observed transitions \((s_t , a_t, r_t, s_{t+1})\) are stored in a replay buffer.
At each step, a minibatch of \(N\) transitions is uniformly sampled from the buffer. The parameters of the critic network are then optimized using Adam optimization to minimize the loss given as:
\begin{equation}
\label{eq:onesteploss}
L(\theta^Q) = \frac{1}{N} \sum_{i=1}^N ( y_i - Q(s_i, a_i | \theta^Q) )^2
\end{equation}
\begin{equation}
\label{eq:onesteptarget}
y_i = r_i + \gamma Q'(s_{i+1}, \mu'(s_{i+1} | \theta^{\mu'}) | \theta^{Q'})
\end{equation}
where $y_i$ is the one-step target with the discount factor \(\gamma\). Here, \(Q'(s,a|\theta^{Q'})\) and \(\mu'(s|\theta^{\mu'})\) are the target networks associated with \(Q(s,a|\theta^Q)\) and \(\mu(s|\theta^\mu)\). Their parameters are updated at each step using soft updates, i.e. \(\theta' \leftarrow \tau \theta + (1-\tau) \theta'\) with \(\tau \ll 1\).
To update the parameters of the actor network, a step proportional to the sampled gradient of the critic network with respect to the action is taken, which is given by:
\begin{equation}
\nabla_{\theta^\mu}J \approx \frac{1}{N} \sum_{i=1}^N \nabla_a Q(s,a|\theta^Q)|_{s=s_i,a=\mu(s_i)} \nabla_{\theta_\mu} \mu(s|\theta^\mu)|_{s=s_i}.
\end{equation}
\cite{DBLP:journals/corr/LillicrapHPHETS15} evaluated DDPG on a car racing problem. They reported that, for both using low-dimensional data and using pixels as input, some replicas learned reasonable policies, while others did not. An open source implementation of DDPG~\footnote{https://yanpanlau.github.io/2016/10/11/Torcs-Keras.html} replicates those results to learn steering, or break. Instead, we introduce numerous modifications to the algorithm that make it possible to learn reasonable policies for racing and outperforming well-known baselines.

\textbf{Window sampling.} 
In partially observable environments, accessing a single state does not reveal the full underlying state of the environment at each time step. 
Window sampling provides the agent additional information by feeding a window of the last \(w\) states to the actor and the critic network. 
%\red{During learning, transitions are sampled from the replay buffer together with the previous \(w-1\) states for each transition.}

\textbf{Long short term memory.}
Another natural approach to include knowledge from past experiences is to make use of recurrent neural networks, 
%While in the window sampling approach the agent receives information only about the \(w\) last states, recurrent neural networks 
which are able to remember information for an arbitrary number of time steps \cite{Hochreiter:1997:LSM:1246443.1246450}. For example, LSTM units can be added to the actor and the critic network.

\textbf{Multi-step targets.}
For updating the critic function, a one-step target is used in standard DDPG. 
%\red{i.e. the target consists of the reward obtained when taking action \(a_t\) in state \(s_t\) and the discounted target Q-value for state \(s_{t+1}\) and the action given by the target Q-network for that state.} 
Multi-step targets \cite{Mnih:2016:AMD:3045390.3045594,DBLP:journals/corr/VecerikHSWPPHRL17} incorporate the next \(n\) rewards obtained along the trajectory starting from state \(s_t\) and following a policy close to the current policy \(\mu(s|\theta^\mu)\) at time step \(t\). The one-step target \(y_i\) (in Eq.~\ref{eq:onesteptarget}) is replaced by:
\begin{equation}
y_i^{(n)} = \sum_{k=0}^{n-1} \gamma^k r_{i+k} + \gamma^n Q'(s_{i+n}, \mu'(s_{i+n} | \theta^{\mu'}) | \theta^{Q'})
\end{equation}
%\red{During learning, transitions are sampled from the replay buffer together with the subsequent \(n-1\) rewards for each transition.}

\textbf{Prioritized experience replay.}
In standard DDPG, transitions are sampled uniformly from the replay buffer at each step. Prioritized experience replay (PER) \cite{DBLP:journals/corr/SchaulQAS15} attempts to make learning more efficient by sampling more frequently transitions that are more important for learning. The probability of sampling a particular transition \(i\) from the replay buffer is given by 
$P(i) = \frac{p_i^\alpha}{\sum_k p_k^\alpha}$,
where \(p_i\) is the transition's priority. The sum in the denominator runs over all transitions in the buffer. Similar to the implementation outlined in \cite{DBLP:journals/corr/VecerikHSWPPHRL17}, we use $p_i = \delta_i^2 + \lambda_3 |\nabla_a Q(s_i,a_i|\theta^Q)|^2 + \epsilon$. 
Here, \(\delta_i\) is the temporal-difference error calculated for the transition when it was sampled the last time, the second term represents the transition's contribution to the actor loss, \(\lambda_3\) is used to weight the two contributions, and \(\epsilon\) is a small positive constant ensuring all transitions are sampled with some probability. 
%\red{New transitions are inserted into the replay buffer with maximum priority to make sure they are sampled at least once. To account for the bias introduced by changing the distribution, updates to the critic network are weighted with importance sampling weights, $w_i = \left( \frac{1}{N} \frac{1}{P(i)} \right)^\beta$, where \(\beta=1\) corresponds to a full compensation for the non-uniform probabilities \(P(i)\).}
 
 \begin{table*}[ht]
    \centering
    \begin{varwidth}{\linewidth}
        \centering
        \setlength{\tabcolsep}{0.35em}
        \begin{footnotesize}
            \begin{tabular}{lp{11cm}}
                \toprule
                Notation & Description \\
                \midrule
                \(\theta\) & Angle between the car direction and the direction of the track axis or racing line. \\
                track & Vector of 19 range finder sensors:  each sensor returns the distance between the track edge and the car within a range of  200\,m. \\
                trackPos & Distance between the car and the track axis or racing line. \\
                \(V_x\) & Speed of the car along its longitudinal axis. \\
                \(V_y\) & Speed of the car along its transverse axis. \\
                \(V_z\) & Speed of the car along its \(z\)-axis. \\
                \(\vec{\omega}\) & Vector of 4 sensors representing the rotation speed of the wheels. \\
                \(f_{\text{rot}}\) & Number of rotations per minute of the car engine. \\
                \(LAC\) & Look ahead curvature. Vector of 4 curvature measurements from the racing line at 20, 40, 60 and 80 meters ahead. The curvature is recorded from a previous slow lap.\\     
                \bottomrule
            \end{tabular}
        \end{footnotesize}
        %\caption[]{Telemetry features used as input \cite{DBLP:journals/corr/abs-1304-1672}.}
        %\label{table:features}
    \end{varwidth}%
    \begin{minipage}{0.19\linewidth}
        \centering
        \includegraphics[height=1.3cm]{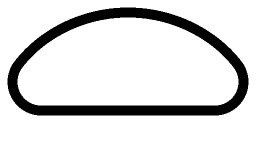}
        \includegraphics[height=1.3cm]{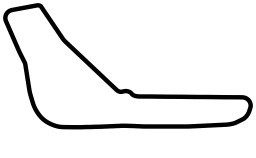}
        \includegraphics[height=1.6cm]{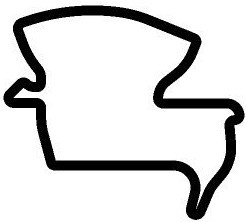}
        %\captionof{figure}{Tracks used for evaluation \cite{TORCS}.}
        %\label{fig:tracks}
    \end{minipage}
    \caption[]{\emph{Left:} Telemetry features used as input \cite{DBLP:journals/corr/abs-1304-1672}. \emph{Right:} Tracks used for evaluation \cite{TORCS}: Michigan (Top), Forza (Middle), Aalborg (Bottom). Note that LAC is not used in the 1st and 2nd part of Study 1.}
    \label{table:telem_tracks}
 \end{table*}
\subsection{Extensions of RL Algorithms}

To ensure that DDPG variants work on racing tracks, we introduced: a method to reduce the exploration of continuous action spaces with two mutually exclusive outputs, a modification of the training objective for episode terminations, and a variation of the critic network.

\textbf{Brake exploration}
The most important actions to control a racing car are steering, accelerating, and braking. Learning how to steer is rather straightforward, but the complex interplay between brake and throttle is very challenging from the exploration perspective.
Following \cite{DBLP:journals/corr/LillicrapHPHETS15}, we use an Ornstein-Uhlenbeck (OU) process \cite{PhysRev.36.823}, which outputs temporally correlated values centered around a given mean, for noise generation. This allows for temporally correlated exploration. The values generated by the OU process are attenuated proportional to a parameter $\epsilon'$, where $\epsilon'$ is set to 1.0 in the beginning and is annealed to zero at the end of exploration phase.
We add noise to three action dimensions independently, but with a probability of 0.1. We stochastically add a stronger noise to the brake while simultaneously lowering throttle by a factor of $(1-\epsilon')$. This guides the exploration to not press the throttle and brake simultaneously, reducing the exploration space drastically. 

\textbf{Episode termination}
%We consider three types of episode termination: i) when reaching a certain number of steps or ii) when the car is out of track, moves backwards, or iii) its progress along its longitudinal axis is slow. For the latter (iii), the agent receives a reward of \(-1\) at termination. In ii), the target (Equation \ref{eq:onesteptarget}) for the end-of-episode transition is replaced by \(y_i = r_i\). When terminating because the number of time steps is reached (i), Equation \ref{eq:onesteptarget} is used as target for the end-of-episode transition as for non-terminal time steps. This is done as the occurrence of a good termination does not depend on decisions of the agent, but only on the definition of the maximum number of steps. Changing the target to \(y_i = r_i\) (and thus to a lower value) in this case might lead to the agent to an (unintended) adaption of the weights. This small modification is essential, especially when episodes are terminated after only a few steps as is done when training to optimize single corners. The learning curves can be found in the supplementary material.
Episodes are terminated when a certain number of steps is reached, the car is out of track, moves backwards, or its progress along its longitudinal axis is slow. In all these cases except for the last, the agent receives a reward of \(-1\) at termination. In all cases, the target (Equation \ref{eq:onesteptarget}) for the end-of-episode transition is replaced by \(y_i = r_i\). The problem with this approach is that when terminating an episode due to reaching the maximum number of steps, the target function would also be \(y_i = r_i\) for this end-of-episode transition, despite it being a good episode termination. Instead, to prevent an unintended adaption of the weights, in this case we use the same target (Equation~\ref{eq:onesteptarget}) as for non-terminal transitions.
\[y_i=\begin{cases}
r_i & \parbox{2.4cm}{\textit{premature end of episode}} \\
r_i + \gamma Q'(s_{i+1}, \mu'(s_{i+1} | \theta^{\mu'}) | \theta^{Q'}) & \parbox{2.4cm}{$i=max\_steps$ \textbf{or} $normal\ step$}
\end{cases}\]

%\cite{Hochreiter:1997:LSM:1246443.1246450} 
\textbf{LSTM critic network}
Initially, we added LSTM units to both the actor and the critic network, but this lead to an unstable behavior of the actor. Thus, following the DQN approach for discrete action spaces in \cite{DBLP:journals/corr/HausknechtS15}, we kept the LSTM only in the critic. 
This is expected to improve the performance of the algorithm, as a good approximation of the Q value function is the basis for learning a good policy.  
The LSTM layer is placed after the concatenation of the state and the action stream. The input to the critic  is a window of the last \(w\) states and actions. 
%\red{During learning, transitions are sampled from the replay buffer together with the previous \(w-1\) states and actions for each transition.}

\section{Experiments}

%We propose two studies to answer our research questions. Study 1 on \emph{learning to drive} concerns RQ1 and RQ2. It compares the performance of state-of-the-art variants of DDPG to assess the efficiency in learning to drive solely with telemetry data. Study 2 on \emph{driving in new scenarios} concentrates on RQ3. It repeats the training of the best performing methods from Study 1 on various tracks to determine which driver model manages not only to drive the track it was trained on, but is also able to generalize to unknown tracks.

%\subsection{Environment}

We evaluate the different algorithms on the open-source simulator TORCS (The Open Racing Car Simulator) \cite{TORCS}. TORCS is used for research in RL and autonomous driving, utilizing either images or low-dimensional features as input, e.g.~\cite{DBLP:journals/corr/LillicrapHPHETS15}. 
%\red{It utilizes simplified physics models to simulate basic vehicular dynamics.} 
At each time step, the agent receives detailed information about the state of the environment. However, as many parts of the simulation are not directly accessible to the agent, the environment is partially observable even if only a single car is on the track. The interaction with the environment takes place in discrete time steps with a spacing of 200\,ms. The input to the algorithms consists of vehicle telemetry data. We carefully selected the features in Table~\ref{table:telem_tracks}. 
The reward is calculated as: 
\begin{equation}
r = V_x (\cos(\theta) - \sin(\theta) - |\text{distance to track axis}|) 
\end{equation}
where $\theta$ is the angle between the car direction and racing line.
%This function maximizes the velocity in the direction of the track axis, minimizes the velocity transverse to the track axis, and penalizes the agent for not exactly following the track axis. 

%We compare three approaches: i) the track axis is the middle line of the track. However, this reward function is unlikely to lead to an optimal driving behavior, as the middle of the track is not the optimal racing line. ii) a racing line from the Tita bot. The algorithm optimizes its path taking the racing line as a reference. This reduces the exploration space significantly. iii) same as ii) but adding the $LAC$ to the input state $s_t$.

%In the first two experiments, we compare three approaches:  i) the track axis is the middle line of the track (MOT), ii) a racing line from the Tita bot and iii) the look ahead curvature (LAC) to the input of $s_t$.
%The algorithm optimizes its path taking the racing line as a reference. 
%This drastically reduces the exploration space.  We hypothesise that when using the middle of the track as racing line, the reward function is unlikely to lead to an optimal driving behavior, as the middle of the track is not the optimal racing line. 

We compare three approaches: i) the middle line of the track is considered to be the track axis, which is unlikely to lead to an optimal driving behavior, as the middle of the track is not the optimal racing line. ii) a racing line from the Tita bot. The algorithm optimizes its path taking the racing line as a loose reference which then it improves by generating a new. This reduces the exploration space significantly. iii) same as ii) but adding the $LAC$ to the input state $s_t$.

\textit{Damage} is a quantity calculated by TORCS each time the car hits a wall and is proportional with the impact of the hit on the car. We consider the negative of the damage magnitude in the reward function and we report the damage of a model as the cumulative sum over the entire testing phase.

% TODO: ADD INFORMATION ABOUT THE CAR USED
The experiments use three tracks considered by \cite{Cardamone:2010:SPRLA} with increasing complexity (see Table~\ref{table:telem_tracks}). 
The Michigan Speedway is a semi-oval track that can be driven without using the brake. Forza and Aalborg are far more complex and cannot be driven without braking. Especially Aalborg is rather technical, with sharp turns that have to be taken at lower speed as well as fast segments.

\subsection{Algorithms}

%\emph{Algorithms.} 
We considered four algorithms: DDPG, LSTM, MS and PER and each of them was tested with $2-3$ variation of hyperparamaters. 
We used four variations of DDPG:
%\begin{enumerate}
%\item
\emph{\footnotesize WIN1, WIN4, WIN8} use window sampling with window sizes 1, 4, and 8 (WIN1 is standard DDPG).
%\item 
\emph{\footnotesize LSTM4, LSTM8} utilize an LSTM critic network with window sizes 4 and 8.
%\item 
\emph{\footnotesize MS2, MS3, MS4} uses multi-step targets with 2, 3, and 4 steps.
%\item 
\emph{\footnotesize PER40k, PER1M} utilize PER with buffer sizes \(4\times10^4\) and \(10^6\).

\subsection{Study 1: Learning to Drive}

%We propose two studies to answer our research questions. 
This study addresses RQ1 by examining whether RL models can drive a racing car from telemetry data and then comparatively evaluating the performance across different models. 

\textbf{Procedure.}
This study was split in three parts. To reduce the training time for hyperparameter selection, part1 uses a simple track, Michigan and trained for 500 episodes. 
The learned model was \emph{tested} without exploration ($\epsilon=0$) on the same track the results were used to select the hyperparametrs for each of the four used algorithms.
In part2, the four algorithms (with selected hyperparameters) and tested on a technically more complex track, Aalborg, using both MOT and Tita bot as racing line references. With the selected best algorithm of part2, in part3 we evaluated the impact of adding future information of the track. We did so by adding the look ahead curvature (LAC) and training in all three tracks. Part 2 and 3 were trained for 7000 episodes. The results were used to select the best algorithm/model suited for the task of autonomous racing.

\begin{comment}
The study involves three parts: the first part uses the Michigan track, as it is simple and the number of episodes required to learn a good policy is rather low, which makes it ideal candidate for selecting hyperparamaters for each algorithm. %In this case, the middle of the track (MOT) is used as racing line.
%and limited the action space to steering and throttle (no brake)}. 
%For each algorithm, 10 training runs were performed. 
After a training run, the learned model was \emph{tested} without exploration ($\epsilon=0$) on the same track. 
%To assess the learning, we report the testing episode reward during training, the overall best lap-time, average best lap-time, and average damage.
Lap-time has highest priority in car racing and is commonly used to compare the performance of different models. Hence, we considered the average best lap-time achieved in the first part of the study to select the hyperparemeters algorithms for the second part, in order to reduce the training time. 
In the second part, the selected algorithms were trained and tested on a technically more complex track, Aalborg, using both MOT and Tita bot as racing line references. In the third part, we evaluate the impact of adding future information of the track to the model by adding the look ahead curvature (LAC).
\end{comment}

\textbf{Results.} 
Lap time is the most important measure in racing and we will utilise it to compare the performance of different models. 
Table~\ref{table:michigan} shows best (bLT) and average (aLT) lap-times in both Michigan and Aalborg as well as baseline approaches: Tita (heuristic state-of-the-art bot) and Genetic \cite{Cardamone:2010:SPRLA}. 
The results (aLT) of Michigan track were used to select the best hyperparameters for each algorithm.
Thus in turn, WIN4, MS4, PER1M and LSTM4 were selected to be used for Aarlborg track. 
As shown in Table~\ref{table:michigan}, our models outperform the baseline bots by a large margin (bLT: WIN1 $26.75$ vs Tita $28.57$) on the simple track. On the complex track, only models trained with racing line (RC) are faster (LT: PER1M $67.17$ vs Tita $68.11$). 
The results also show that PER1M trained with racing line achieves the best lap-time and thus it is selected to compare the effect of adding the LAC. Table~\ref{table:curvature} shows the results of adding the LAC to the state (PER1M). This addition gives the best result.

%Figure~\ref{fig:michigan} illustrates the training performance in terms of average episode reward for all algorithms on the Michigan track. 

%when learning to control steering and throttle on the Michigan track.

\begin{table}[t]
    \centering
    \footnotesize
        \begin{tabular}{p{1.2cm}p{0.8cm}p{0.8cm}|p{0.7cm}p{0.7cm}|p{0.7cm}p{0.7cm}}
            %  \toprule
            & \multicolumn{2}{p{2.05cm}|}{Part I: Michigan} & \multicolumn{4}{c}{Part II: Aarlborg}\\ \cline{2-7}
            & \multicolumn{4}{c|}{MOT Reference} & \multicolumn{2}{c}{RC Reference}\\ \cline{2-7} 
            &{bLT}  &{aLT}  & {bLT}    & {aLT}    & {bLT}    & {aLT}   \\ \midrule
            WIN1        & \textbf{26.75} & 27.16          &{-} &{-} &{-} &{-} \\
            WIN4        & 26.77         & \textbf{26.96} &  74.75       &  75.73 &   67.56   &   \textbf{69.89}     \\
            WIN8        & 26.75         & 35.61          &{-} &{-} &{-} &{-}\\  \midrule
            MS2         & 26.79         & 30.33          &{-} &{-} &{-} &{-}\\  
            MS3         & 26.80         & 27.13         &{-} &{-} &{-} &{-}\\   
            MS4         & 26.83         & 27.04         &  77.68       &  78.53 &    69.94    &   77.01     \\  \midrule
            PER40k      & 26.84         & 34.85         &{-} &{-} &{-} &{-}\\   
            PER1M       & 27.06         & 32.30         &  71.76       &  75.22 &    \textbf{67.17}  &   70.11           \\ \midrule
            LSTM4       & 27.35         & 27.52         &{85.87} &{90.05} &{-} &{-}\\
            LSTM8       & 27.50         & 27.85         &{-} &{-} &{-} &{-}\\   \midrule \midrule
            Tita        & 28.57         & {-}           &  {-}           &  {-}     &    68.11  &   {-}                \\
            Genetic     & 33.86         & {-}           &  {-}           &  {-}     &    69.92  &   {-}                   \\
            \bottomrule
        \end{tabular}
        \caption{Testing results over 10 runs in Michigan and over 5 runs in Aalborg. Best lap-time (\textbf{bLT}), average best lap-time (\textbf{aLT}) (in seconds) for models with 0 damage. The Aalborg track was trained/tested using both middle of the line (\textbf{MOT}) and racing line from Tita (\textbf{RC}) bot as racing line references.}
    \label{table:michigan}
\end{table}

\begin{comment}
\begin{figure}[t]
    \centering
    \includegraphics[height=5.4cm]{ddpg_all_val.png}
    \caption{Testing episode reward during first experiment. The reward is averaged over 10 training runs, smoothed with a moving average over 20 episodes.}
    \label{fig:michigan}
\end{figure}
\end{comment}

\subsection{Study 2: Driving in New Scenarios}

Study 2 on \emph{driving in new scenarios} addresses RQ2 and investigates how the driver models perform in unseen tracks.

%Study 2 consists of two \emph{competitions} on the three tracks in Figure~\ref{fig:tracks}: in i)~we put algorithms trained on the Michigan track to race on each track and in ii)~we put algorithms trained on the Aalborg track to race on each track (without additional learning). 

\textbf{Procedure.}
For this study, we use different variants of PER1M %and WIN4 
(MOT, RC, RC+LAC) as shown in Table~\ref{table:traintest}. The models are trained in one track (Michigan or Aalborg) and tested in two other unseen tracks. While training, as the model improves, a test is performed in the other two unseen tracks. We choose as final model, referred to as \textit{general model}, the one that performs with the best lap time in the training track but that is also able to finish the unseen tracks.

\begin{figure}[hbt]
    \centering
    \includegraphics[height=2.5cm]{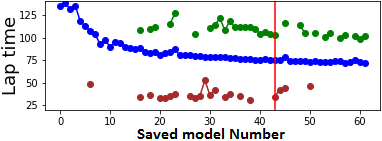}
    \caption[]{Fastest laps for all PER1M models trained on Aalborg and tested on Aalborg (Blue), Michigan (Brown), and Forza (Green).}
    \label{fig:generalization}    
\end{figure}

\textbf{Results.}
First, the models trained in Michigan did not finish the unseen tracks (Aalborg and Forza). We attribute this to the simplicity of the Michigan track (no sharp corners) offering little exposure to complex manoeuvres and leading quickly to overfitting.
Table~\ref{table:traintest} shows the results of the general models trained in Aalborg and tested in unseen tracks. These models were able to drive in unseen tracks. As expected, their performance in unseen tracks compared to the performance of models trained and tested in the same track (see Table~\ref{table:curvature}) is significantly worse.
Figure~\ref{fig:generalization} illustrates lap-times achieved on each track for all PER1M models learned on Aalborg. The vertical red line indicates the \emph{general} model. With more training, the performance improvements in Forza (test) are consistent with those in Aalborg (training), i.e., the more the model trained in Aalborg, the better it performed in Forza. But, these improvements were not reflected in the Michigan track, which we think is due to the very basic shape of the Michigan track (no steep curves).

\begin{table}[hbt]
	\centering
	\footnotesize
	\setlength{\tabcolsep}{0.28em}
	\begin{tabular}{l|ccc}
		\toprule

		           &Forza    &Michigan   &Aalborg 					      \\
		\midrule
		%Aalborg       &&& \\	
		%\midrule	
		Tita       &77.39   &28.57        &68.11  			  \\
		PER1M MOT        &-     &27.060    &       71.758        \\
		PER1M RC Reference        &75.306  &26.906	 &67.27\\  
		PER1M RC Reference + LAC    &\textbf{73.018}  &\textbf{26.278} &\textbf{63.35}	\\  		
		\bottomrule
	\end{tabular}
	    \caption{Lap times (best) of a comparison between the best models of the considered different approaches in Study 1. \textbf{Tita} vs \textbf{PER1M} with the reward function set to follow the middle of the track (\textbf{MOT}) vs bot racing line vs bot racing line + look ahead curvature (\textbf{LAC}).}
	    \label{table:curvature}
\end{table}
\begin{table}[hbt]
    \centering
    \begin{varwidth}{\linewidth}
    \footnotesize
    \setlength{\tabcolsep}{0.28em}
    \begin{tabular}{ccc|ccc}
                        & \multicolumn{2}{c}{Best} & \multicolumn{3}{c}{General}       \\ \toprule
                        & {bLT}    & {aLT}   & {Aalborg} & {Michigan}    & {Forza}      \\ \midrule
        PER1M MOT       & 71.76     & 75.22   & 74.84       &    34.65      & 103.09    \\
        PER1M RC Ref.       & 67.17     & 70.11   & 71.26       &    33.67      & 108.69    \\        
        PER1M RC Ref. + LAC         & 63.35     & 65.14   & 70.616  & 34.07 & 107.15     \\     
        %WIN4 MOT        & 74.75     & 75.73   & 82.68       &    35.93      & DNF       \\        
        %WIN4 RC Ref.         & 67.60     & 69.89   & 80.63       &    40.21      & 124.66    \\ 
        % WIN4 RC Ref. + LAC         & ADD     & ADD   & ADD      &    ADD      & ADD  -> dind' finish  \\            
\bottomrule  
    \end{tabular}
    \end{varwidth}%
     \caption[]{Fastest models trained on Aalborg compared with \emph{general} models that complete all tracks (Testing). DNF: no model finished.}
    \label{table:traintest}
\end{table}

\section{Impact of the Contributions}

\subsection{Brake Exploration}

\begin{figure}[hbt]
	\centering
	\includegraphics[width=0.4\textwidth]{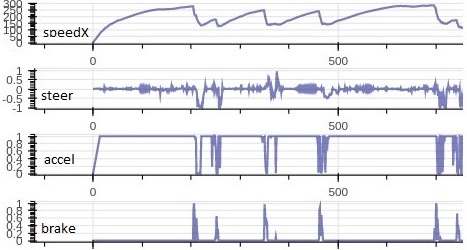}
	\caption{Outputs (steer, acceleration, brake) of a model trained with brake exploration and speed along the car's longitudinal axis.}
	\label{fig:pedal_brake_cropped}
\end{figure}

Figure~\ref{fig:pedal_brake_cropped} depicts the outputs of a model trained with the brake exploration scheme proposed in Section~3.2. Using this exploration scheme, the model is able to press the throttle while completely releasing the brake and vice versa.
This shows that the approach is capable of reducing the exploration space. We also observe that after braking and releasing the pedal, the model waits for some time until it starts steering. This is a common practice among professional racing drivers to avoid over-steering. The algorithm learned this behaviour by itself.

\subsection{Episode Termination}
\begin{figure}[hbt]
	\centering
	\includegraphics[height=4.0cm]{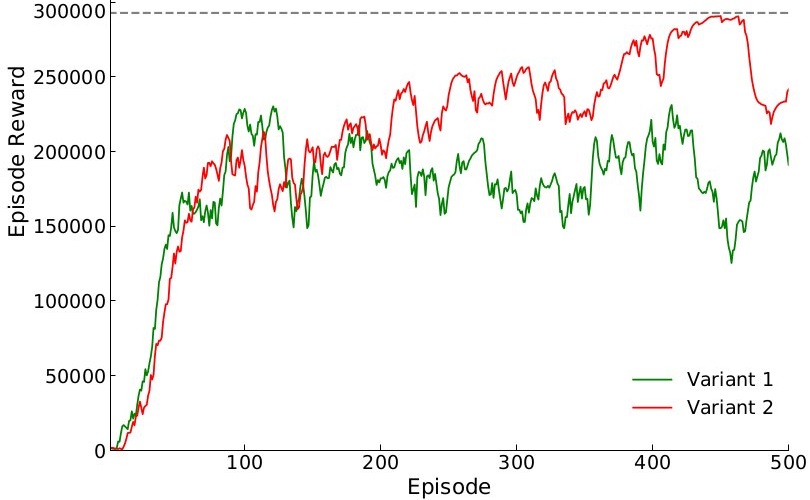}
	\caption{Training episode reward for normal end-of-episode (green) and \textbf{AT} (red) transitions. The reward is averaged over 10 training runs with different random seeds, smoothed with a moving average over 5 episodes.\textbf{WIN1} with episode terminations caused by reaching the maximum number of steps  }
	\label{fig:ddpg_dones}
\end{figure}

We compared two different settings for good episode terminations (terminations caused by reaching the maximum number of steps).
First, we set the target to \(y_i = r_i\) for the corresponding end-of-episode transitions (variant 1). Second, we set the target to \(y_i = r_i + \gamma Q'(s_{i+1}, \mu'(s_{i+1} | \theta^{\mu'}) | \theta^{Q'})\) as for non-terminal transitions (variant 2) and we refer to it as adopted target (\textbf{AT}). Figure~\ref{fig:ddpg_dones} shows the training performance in terms of average episode reward for the WIN1 algorithm for both variants when learning to control steering and pedal on the Michigan track. It can be seen that a higher maximum is reached for variant 2. This small modification is essential, especially when episodes are terminated after only a few steps as is done when training to optimize single corners.

\section{Discussion and Outlook}
% Results Study 1

We presented a comprehensive study applying RL methods to self-driving racing cars relying only on car telemetry data. Our goal was to investigate the capabilities of RL to drive from telemetry data, because simulated environments can produce telemetry data at high rates, which makes the use case practical. Besides, the telemetry data describes the physics vehicle dynamics, which is important since we intended to drive the car at its physical limits.

The results of Study 1 supports \emph{RQ1}: Deep RL algorithms effectively learn to drive fast from telemetry data and obtain models better than state-of-the-art handcrafted models. The results also showed that PER1M (prioritized experience replay with a replay buffer size of 1M samples) was the best performing algorithm in a complex track. Most importantly, the results evidenced that our proposed look ahead of the curve approach (LAC) improves the performance of models in the self racing scenarios. 
To the best of our knowledge, these results constitute a contribution to the line optimization problem using reinforcement learning, and our self-thought models outperform the baseline open-source bots.
Our findings also show that the resulting models are able to work with any racing line, making them suitable for street cars where the problem is typically to follow a given trajectory. 
Note that contrary to bots, our solution does not blindly follow a given racing line, instead, it takes it as loose guide and then it improves it to generate a new optimal racing line.

%\emph{RQ2}: Results from Study 1 evidence the interesting observation that algorithms with a good learning response in training not always deliver the fastest models in the end. 

% Results Study 2
Second, we studied how the learned models perform on unseen tracks. In Study 2, we trained models on a simple and difficult track and compared their ability to drive on unknown tracks. The results show that models trained in reasonably complex tracks (Aalborg) can generalise relatively well in unseen (untrained) simple (Michigan) or more complex tracks (Forza). However, such models underperformed in terms of lap time compared with models trained and tested on the same track. It is interesting that such performance behaviour resembles the behaviour of professional human pilot drivers. Even though they are experts in driving and have acquired a lot of driving skills over the years, expert drivers still practice a lot for the competing track before the competition. This way they can memorise the landmarks and car dynamics before achieving their full performance. %Additionally, when trained in Michigan track, which is a very simple track, the models were not able to generalise. However, this is expected when considering that Michigan is a semi-oval track that can be driven without using the brake, has no sharp corners and there are only left corners.

Similar to human pilots, models trained in Aalborg in Study 2, did acquire driving skills, but in order to achieve the best performance, they would need to continue training for each specific track. Also, similar to human pilots that learn the track by heart, when assisting the model with the information about the curvature using our proposed look ahead of curve (LAC) method, the performance of the model improves. 

Thus, in the future work, we will put more effort and further investigate means of generating such a general model which would achieve relatively high performance in unseen tracks and see how further training it for specific tracks, will affect the performance. This would improve training time. One way to make such general models robust and learn faster would be to use pre-defined maneuvers to train the model.
Additionally, we plan to investigate whether we could train models without using any reference line during the training. 
In this work, we utilised TORCS as a simulation environment which was sufficient to evaluate our hypotheses and to demonstrate the capabilities of our approach, while still being fast to train/test and iterate. %Considering our good results, in future work, we will move towards more sophisticated industrial simulation environments which are considered to be the golden industry standards.
Yet, professional telemetry systems provide in real-time a wealth of information otherwise inaccessible, such as engine temperature variations, damper displacement, tire saturation. In future work, we intend to move towards golden industry standards in simulated environments and make use of more sophisticated telemetry information. Finally, we intend to investigate the interactions with using images as a baseline for comparison but more importantly as complementary channels.

\section*{Acknowledgement}
\footnotesize
This research was partially funded by AVL GmbH and Know-Center GmbH. Know-Center is funded within the Austrian COMET Program - Competence Centers for Excellent Technologies - under the auspices of the Austrian Federal Ministry of Transport, Innovation and Technology, the Austrian Federal Ministry of Economy, Family and Youth and by the State of Styria. COMET is managed by the Austrian Research Promotion Agency FFG.

\bibliographystyle{IEEEtranS}
\small
\bibliography{references}

%\clearpage
%\thispagestyle{empty}
%\input{extra-content.tex}

\end{document}